\documentclass[default,iicol]{sn-jnl}

\usepackage{fancyhdr}
\usepackage{xcolor} 
\usepackage{algorithm}
\usepackage{algpseudocode}
\usepackage{natbib}
\usepackage{eso-pic} 
\usepackage{forloop}
\usepackage{url}
\usepackage{microtype}
\usepackage{graphicx}
\usepackage{subfigure}
\usepackage{booktabs} 
\usepackage{esvect}
\usepackage{mathtools, nccmath}
\usepackage{gensymb}
\usepackage{hyperref}
\usepackage{stackengine,scalerel}
\usepackage{calc}
\newlength\shlength
\newcommand\xshlongvec[2][0]{\ThisStyle{\setlength\shlength{#1\LMpt}%
  \stackengine{-5.6\LMpt}{$\SavedStyle#2$}{\smash{$\kern\shlength%
    \stackengine{\dimexpr 1.3pt+6.25\LMpt}{$\SavedStyle\mathchar"017E$}%
      {\rule{\widthof{$\SavedStyle#2$}}{\dimexpr.1pt+.5\LMpt}\kern.4\LMpt}{O}{r}{F}{F}{L}\kern-\shlength$}}%
      {O}{c}{F}{T}{S}}}
\usepackage{amsmath}
\usepackage{amssymb}
\usepackage{mathtools}
\usepackage{amsthm}

\theoremstyle{plain}

\theoremstyle{definition}

\theoremstyle{remark}

\usepackage{caption}
\jyear{2022}%

\begin{document}

\title[MixMOBO]{Bayesian Optimization For Multi-Objective Mixed-Variable Problems}


\author*[1]{\fnm{Haris Moazam} \sur{Sheikh}}\email{harissheikh@berkeley.edu}

\author[1]{\fnm{Philip S.} \sur{Marcus}}\email{pmarcus@me.berkeley.edu}


\affil[1]{\orgdiv{CFD Lab, Department of Mechanical Engineering}, \orgname{University of California, Berkeley}, \orgaddress{ \state{CA}, \country{USA}}}




\abstract{Optimizing multiple, non-preferential objectives for mixed-variable, expensive black-box problems is important in many areas of engineering and science. The expensive, noisy, black-box nature of these problems makes them ideal candidates for Bayesian optimization (BO). Mixed-variable and multi-objective problems, however, are a challenge due to BO's underlying smooth Gaussian process surrogate model. Current multi-objective BO algorithms cannot deal with mixed-variable problems. We present MixMOBO, the first mixed-variable, multi-objective Bayesian optimization framework for such problems. Using MixMOBO, optimal Pareto-fronts for multi-objective, mixed-variable design spaces can be found efficiently while ensuring diverse solutions. The method is sufficiently flexible to incorporate different kernels and acquisition functions, including those that were developed for mixed-variable or multi-objective problems by other authors. We also present HedgeMO, a modified Hedge strategy that uses a portfolio of acquisition functions for multi-objective problems. We present a new acquisition function, SMC. Our results show that MixMOBO performs well against other mixed-variable algorithms on synthetic problems.  We apply MixMOBO to the real-world design of an architected material and show that our optimal design, which was experimentally fabricated and validated, has a normalized strain energy density $10^4$ times greater than existing structures.}




\maketitle

\section{Introduction}
\label{sec:2}
Optimization is an inherent part of design for complex physical systems. Often optimization problems are posed as noisy black-box problems subject to constraints, where each function call requires an extremely expensive computation or a physical experiment. Many of these problems require optimizing a mixed-variable design space (combinatorial, discrete, and continuous) for multiple non-preferential objectives. Architected material design \cite{frazier2015, chen2018computational, chen2019stiff, shaw2019computationally, song2019topology,sadvours}, hyper-parameter tuning for machine learning algorithms \cite{snoek2012practical, alphago2018, pmlr-v80-oh18a}, drug design \cite{articledrug, pmlr-v108-korovina20a}, fluid machinery \cite{airfoils,draft,doi:10.1177/0309524X17709732,2021APS..DFDA15004S} and, controller sensor placement \cite{JMLR:v9:krause08a} pose such problems. Due to their cost of evaluation, Bayesian optimization is a natural candidate for their optimization.

Much research has gone into Bayesian optimization for continuous design spaces using Gaussian processes (GP) as a surrogate model and efficiently optimizing this design space with a minimum number of expensive function calls \cite{Mockus1994,rasmussen,brochu2010tutorial}. Despite the success of continuous variable Bayesian optimization strategies, multi-objective and mixed-variable problems remain an area of open research. The inherent continuous nature of GP makes dealing with mixed-variable problems challenging. Finding a Pareto-front for multi-objective problems, and parallelizing function calls for batch updates, Q-batch, also remain challenges in the sequential setting of the BO algorithm. Hedge strategies, which use a portfolio of acquisition functions to reduce the effect of choosing a particular acquisition function, have not been formulated for multi-objective problems.

\subsection{Mixed-Variable BO Algorithms:}
We provide a brief description of the current approaches in recent studies for dealing with mixed variables.

\textbf{One Hot Encoding Approach:} Most BO schemes use Gaussian processes as surrogate models. When dealing with categorical variables, a common method is `one-hot encoding' \cite{golovin}. Popular BO packages, such as GPyOpt and Spearmint \cite{snoek2012practical}, use this strategy. However, this can result in inefficiency when searching the parameter space because the surrogate model is continuous. For categorical variables, this approach also leads to a quick explosion in dimensional space  \cite{ru2020bayesian}.

\textbf{Multi-Armed Bandit (MAB) Approach:} Some studies use the MAB approach when dealing with categorical variables where a surrogate surface for continuous variables is defined for each bandit arm. These strategies can be expensive in terms of the number of samples required \cite{NEURIPS2018_cc709032, nguyen2019bayesian}, and they do not share information across categories. An interesting approach, where coupling is introduced between continuous and categorical variables, is presented in the CoCaBO algorithm \cite{ru2020bayesian}, and it is one of the baselines that we test MixMOBO against.

\textbf{Latent Space Approach:} A latent variable approach has also been proposed to model categorical variables \cite{ qian2008gaussian,zhou2011simple,  zhang2020bayesian, DBLP:journals/corr/abs-2111-01186}. This approach embeds each categorical variable in a $\mathcal{Z}$ latent variable space. However, the embedding is dependent on the kernel chosen, and for small-data settings can be inefficient.

\textbf{Modified Kernel Approach:}
There is a rich collection of studies in which the underlying kernel is modified to work with ordinal or categorical variables. For example, \citet{ru2020bayesian} considers the sum + product kernel; \citet{deshwal2021bayesian} proposes hybrid diffusion kernels, HyBO; and \citet{oh2021mixed} proposes frequency modulated kernels. The BOCS algorithm \cite{baptista2018bayesian} for categorical variables uses a scalable modified acquisition function. \citet{pelamatti2018efficient,oh2019combinatorial,nguyen2019bayesian,Garrido_Merch_n_2020} all use modified kernels to adapt the underlying surrogate surface.
Our approach is unique in that any modified kernel can be incorporated into our framework. Currently we use the modified radial basis function (RBF) kernel for modelling the surrogate surface, with our future research focused on using different kernels in our framework.

\textbf{Other Surrogate Models:}
Other surrogate models can be used in place of the  GP to model mixed-variable problems such as random forests, an approach used by SMAC3 \cite{lindauer2021smac3} or tree based estimators, used in the Tree-Parzen Estimator (TPE) \cite{pmlr-v28-bergstra13}. \citet{2020} considers a linear model with cross-product features. BORE \cite{tiao2021bore} leverages the connection to density ratio estimation.

\subsection{Multi-Objective BO Algorithms:}
Multi-objective Bayesian optimization (MOBO) has been the subject of some recent studies. BoTorch \cite{balandat2020botorch}, the popular BO framework, uses the EHVI and ParEGO based on the works of \citet{1688440} and \citet{daulton2020differentiable,daulton2021multiobjective}. Hyper-volume improvement is the main mechanism used to ensure diversity in generations. `$Q$-batch' parallel settings of the above two acquisition functions  use hyper-volume improvement and the previously selected point in the same batch to choose the next set of points. For most single-objective BO algorithms with parallel batch selection, the next batch of test points is selected by adding the `fantasy' cost-function evaluation, usually the predicted mean, to the previously selected test point within that batch. However, this commonly used method often leads to overly confident test point selection, and the surrogate surface then needs to be optimized, and sometimes refitted $Q$ times. Using a genetic algorithm (GA), we can select a `$Q$-batch' of points with a single optimization of the surrogate surface from the GA generation.

\citet{suzuki2020multiobjective} provide an interesting Pareto-frontier entropy method as an acquisition function, and \citet{10.1115/1.4046508} use Pareto-frontier heuristics to formulate new acquisition functions. Their approaches were not extended to mixed-variable problems. 

\subsection{Hedge Strategies}
Hedge algorithms have proven to be efficient in dealing with a diverse set of problems by using a portfolio of acquisition functions. `GP-Hedge', introduced by \citet{brochu2011portfolio} is a well-known and efficient algorithm. However, current Hedge algorithms have not been extended for multi-objective problems and, to the authors' knowledge, there is no existing Hedge strategy implementation that solves such problems.

\section{MixMOBO}
In this paper, we present a Mixed-variable, Multi-Objective Bayesian Optimization (MixMOBO) algorithm, the first generalized framework that can deal with mixed-variable, multi-objective problems in small data setting and  can optimize a noisy black-box function with a small number of function calls. 

Genetic algorithms, such as NSGA-II \cite{996017}, are well known for dealing with mixed-variable spaces and finding an optimal Pareto-frontier. However, these algorithms require a large number of black-box function calls and are not well suited to expensive small-data problems. Our approach is to use a GA to optimize the surrogate model itself and find a Pareto-frontier. Diversification is ensured by the distance metrics used while optimizing the surrogate model. This method allows cheap Q-batch samples from within the GA generation, and also allows the use of commonly used acquisition functions such as Expected Improvement (EI), Probability of Improvement (PI) and Upper Confidence Bound (UCB) \cite{brochu2011portfolio}, which work well for single objective problems. We note here that other metrics can easily be incorporated instead of a distance metric within the GA setting and is one of the areas of our future work. Using a GA on a mixed variable surrogate model in a multi-objective setting allows us to work with modified kernels that were developed for mixed-variable problems in literature. We also present a new acquisition function, `Stochastic Monte-Carlo' (SMC), which performs well for multi-objective categorical variable problems \cite{sadvours}. 

Hedge strategies for Bayesian optimization are efficient for single objective algorithms. We present here our Hedge Multi-Objective (HedgeMO) algorithm, which uses a portfolio of acquisition functions for multi-objective problems and can work with Q-batch updates. It is an extension of GP-Hedge \cite{brochu2011portfolio}, which has regret bounds, and the same  bounds hold for HedgeMO.

We note here that MixMOBO is designed for mixed-variable, multi-objective problems. Although there are algorithms in the literature that can solve problems with a \textit{subset} of these attributes (e.g. mixed-variable single-objective or multi-objective continuous variable problems), no algorithm, to our knowledge, can deal with all of these attributes. In addition, MixMOBO outputs a batch of query points and uses HedgeMO, the first multi-objective hedging strategy. To the authors' knowledge, no existing approaches can achieve all this within a single framework.

In summary, the main contributions of our work are as follows:
\begin{itemize}
    \item We present Mixed-variable, Multi-Objective Bayesian Optimization (MixMOBO), the first algorithm that can deal with mixed-variable, multi-objective problems. The framework uses GA to optimize the acquisition function on a surrogate surface, so it can use modified kernels or surrogate surfaces developed to deal with mixed-variable problems in previous studies. This extends the capabilities of previous approaches in literature to handle mixed-variable and multi-objective problems as well if adopted within our framework, since our framework is agnostic to the underlying GP kernel over mixed-variables.
    \item GA is used to optimize surrogate models, which allows the optimization of multi-objective problems. `$Q$-batch' samples can be extracted in parallel from within the GA generation without sacrificing diversification.
    \item We present a Hedge Multi-Objective (HedgeMO) strategy for multiple objectives for which regret bounds hold. We also present an acquisition function, Stochastic Monte-Carlo (SMC), which performs well for combinatorial multi-objective problems, and use it as part of our HedgeMO portfolio.
    \item We benchmark our algorithm against other mixed-variable algorithms and prove that MixMOBO performs well on test functions. We applied MixMOBO to a practical engineering problem: the design of a new architected meta-material that was optimized to have the maximum possible strain-energy density within the constraints of a design space. The fabrication and testing of this new material showed that is has a normalized strain energy density that is  $10^4$ times greater than existing unblemished microlattice structures in literature.
\end{itemize}

The rest of the paper is organized in the following manner: Section \ref{PS} defines the optimization problem to be solved with MixMOBO. The detailed workings of MixMOBO and HedgeMO are presented in Section \ref{method}. Section \ref{VT} details the validation tests performed on our framework to test its efficiency and comparison to existing algorithms. Our application of MixMOBO for design of architected materials and its results are presented in Section \ref{AM}.

\section{MixMOBO Problem Statement}
\label{PS}
We pose the multi-objective and mixed-variable problem as: 
\begin{equation}
\vec{w}_{opt}= argmax_{\vec{w} \in \mathcal{W}} (\vec{f}(\vec{w}))
\end{equation}
for maximizing the objective. Here $\vec{f}(\vec{w})=[f_1(\vec{w}),f_2(\vec{w}),\ldots$ $,f_k(\vec{w})]$ are the $K$ non-preferential objectives to be maximized, and $\vec{w}$ is a mixed-variable vector, defined as \{$\vec{w} \in \mathcal{W}\} = \{\vec{x} \in \mathcal{X},\vec{y} \in \mathcal{Y},\vec{z} \in \mathcal{Z}\}$. $\vec{x}$ is an $m$-dimensional vector defined over a bounded set $\mathcal{X}\subset \mathbb{R}^m$ representing $m$ continuous variables. Ordinal and categorical variables are defined as $\vec{y}= [y_1,\ldots,y_n]$ and $\vec{z}= [z_1,\ldots,z_o]$, respectively. Each variable $y_j \in \{O_1,\ldots, O_j\}$ takes one of $O_j$ ordinal `levels' (discrete numbers on the real-number line) and each categorical variable takes a value $z_j \in \{C_1,\ldots, C_j\}$ from $C_j$ unordered categories (that cannot, by definition,  be ordered on the real-number line). $\mathcal{Y}$ and $\mathcal{Z}$ are the ordinal and combinatorial spaces respectively. 

Generally, $\{\vec{w}_{opt}\}$ is a set of Pareto-optimal solution vectors i.e.,  vectors that are not Pareto-dominated by any other vector. A vector $\vec{w}$ is Pareto-dominated by $\vec{w}'$, iff $f_{k}(\vec{w}) \leq f_{k}(\vec{w}') ~ \forall ~ k=1,...K$. This $\{\vec{w}_{opt}\}$ is the optimal set found by MixMOBO, details of which are presented in the following section.


\section{Methodology}
\label{method}
\subsection*{Preliminaries}
Single-objective Bayesian optimization is a sequential optimization technique, aimed at finding the global optimum of a single objective noisy black-box function $f$ with minimum number of evaluations of $f$. For every $i^{th}$ iteration, a surrogate model, $g$, is fit over the existing data set $\mathcal{D}=\{(w_1,f(w_1)), \ldots,(w_i,f(w_i))\}$. An acquisition function then determines the next point $\vec{w}_{i+1}$ for evaluation with $f$, balancing exploration and exploitation. Data is appended for the next iteration, $\mathcal{D}=\mathcal{D} \cup (w_{i+1},f(w_{i+1}))$, and the process is repeated until the evaluation budget for $f$ or the global optimum is reached.

Gaussian processes are often used as surrogate models for BO \cite{rasmussen,kpm}. A GP is defined as a  stochastic process such that a linear combination of a finite set of the random variables is a multivariate Gaussian distribution. A GP is uniquely specified by its mean $\mu(\vec{w})$ and covariance function $ker(\vec{w}$, $\vec{w}')$. The GP is a distribution over functions, and $g(\vec{w})$ is a function sampled from this GP:
\begin{equation}
{\vec{g}}(\vec{w}) \sim GP\bigl({{\mu} ( \vec{w} ) },ker(\vec{w},\vec{w}')\bigr).
\label{Does.eq.something}
\end{equation}

Here, $ker(\vec{w},\vec{w}')$ is the covariance between input variables $\vec{w}$ and $\vec{w}'$. Once a GP has been defined, at any ${\vec{w}}$ the GP returns the mean $\mu({\vec{w}})$ and  variance $\sigma({\vec{w}})$. The acquisition function $\mathcal{A}(\vec{g})$, balances exploration and exploitation, and is optimized to find the next optimal point $\vec{w}_{i+1}$. The success of BO comes from the fact that evaluating $\vec{g}$ is much cheaper than evaluating  $\vec{f}$.

\subsection{MixMOBO Approach}

Our Mixed-variable Multi-Objective Bayesian Optimization (MixMOBO) algorithm extends the single-objective, continuous variable BO approach presented in the preceding section, to more generalized optimization problems and is detailed in Algorithm \ref{alg:momvbo}. 

A single noisy GP surrogate surface is fit for multiple objectives, ${\vec{g}}(\vec{w}) \sim GP\bigl({\vec{\mu} ( \vec{w} ) },ker(\vec{w},\vec{w}')\bigr)$. Note that this is different from Eq.~\ref{Does.eq.something}, since the GP would predict mean for multiple objectives. For details on fitting a single GP to multi-objective data, we refer the reader to \citep[Eq. 2.25-2.26]{rasmussen}. For multiple objectives, the response vector, with $n$-data points, is of size $k \times n$. The predicted variance remains the same, but the predicted mean is a $k \times 1$ vector. This is equivalent to fitting $K$ GP surfaces with the same kernel for all of the surfaces, where $K$ is the total number of objectives. All $K$ objectives are assumed to have equal noise levels. Only one set of hyper-parameters needs to be fit over this single surface, rather than fitting $K$ sets of hyper-parameters for $K$ different surfaces; thus, when $K$ is large, the overall computational cost for the algorithm is reduced. Note that we could fit $K$ different GP surfaces, particularly if different noise levels for different objectives is to be considered, with different hyper-parameters to the data to add further flexibility to the fitted surfaces. This idea will be investigated in our future work. We use LOOCV \cite{kpm} for estimating hyper-parameters since we are dealing with small-data problems.

\begin{algorithm}[!ht]
  \caption{ Mixed-variable Multi-Objective Bayesian Optimization (MixMOBO) Algorithm}
  \label{alg:momvbo}
\begin{algorithmic}[1]
  \State {\bfseries Input:} Black-box function $\vec{f}(\vec{w}): {\vec{w} \in \mathcal{W}}$, initial data set size $N\_i$, batch points per epoch $Q$, total epochs $N$, mutation rate $\beta\in [0,1]$
  \State {\bfseries Initialize:} Sample black-box function $\vec{f}$ for  $\mathcal{D}=\left\{\left(\vec{w}_j,\vec{f}(\vec{w}_j)\right)\right\}_{j=1:N\_i}$
\vspace{0.2cm}
  \For{$n=1$ {\bfseries to} $N$}
\vspace{0.15cm}  
  \State Fit a noisy Gaussian process surrogate function ${\vec{g}}(\vec{w}) \sim GP\bigl ({\vec{\mu} ( \vec{w} ) },ker(\vec{w},\vec{w}')\bigr)$ 
  \State For $L$ total acquisition functions, from each $\mathcal{A}^l$ acquisition function, propose $Q$-batch test-points, $\left\{(\vec{u})_n^l\right\}_{1:Q}=\left\{argmax_{\vec{u} \in \mathcal{W}}\mathcal{A}^l\left(\vec{g}\right)\right\}_{1:Q}$ within the constrained search space $\mathcal{W}$ using multi-objective GA
  \State Mutate point $\left\{(\vec{u})_n^l\right\}_{q}$ within the search space $\mathcal{W}$ with probability rate $\beta$ if $L_2$-norm of its difference with any other member in set $\left\{(\vec{u})_n^l\right\}_{1:Q}$ is below tolerance
  \State Select batch of $Q$ points using HedgeMO, $\left\{\vec{w}_n\right\}_{1:Q}=\textit{HedgeMO}\left(\vec{g},\left\{(\vec{u})_{1:n}^{1:L}\right\}_{1:Q},\mathcal{D}\right)$
  \State Evaluate the selected points from the black-box function, $\{\vec{f}(\vec{w}_n)\}_{1:Q}$
  \State Update $\mathcal{D}=\mathcal{D} \cup \left\{\left(\vec{w}_{n},\vec{f}(\vec{w}_{n})\right)\right\}_{1:Q}$
  \vspace{0.15cm} 
  \EndFor
  \vspace{0.2cm}
\State {\bfseries return} Pareto-optimal solution set $\left\{\left(\vec{w}_{opt},\vec{f}(\vec{w}_{opt})\right)\right\}$
\end{algorithmic}
\end{algorithm}

Gaussian processes are defined for continuous variables. For mixed variables, we need to adapt the kernel so that a GP can  be fit over these variables. Cited works in Section \ref{sec:2} dealt with modified kernels that were designed to model mixed variables. Those kernels can be used in the MixMOBO algorithm. For the current study, we use a simple modified squared exponential kernel:
\begin{equation}
     ker({\vec{w}, \vec{w}'}) \equiv \epsilon _f ^2 \,\, exp \left [-\frac{1}{2}\lvert {\vec{w}}, {\vec{w}}'\rvert _{C}^T \,\, \,\, {\underline{\underline{ M}}} \,\, \,\, \lvert {\vec{w}},  \vec{w}'\rvert _{C}\right],
\label{eq:1}
\end{equation}

where $\vec{\theta} = (\{{\underline{\underline{M}}}\},\epsilon _f)$ is a vector containing all the hyper-parameters, $\{{\underline{\underline{M}}}\}=diag(\vec{h})^{-2}$ is the covariance hyper-parameter matrix and $\vec{h}$ is the vector of covariance lengths. The distance metric, $\lvert{\vec{w}}, {\vec{w}}'\rvert_{C}$, is an concatenated vector, with the distance between categorical variables defined to be  the  Hamming distance, and the distance between continuous variables and the distance between  ordinal variables defined to be their Euclidean distances. Noise is added to the diagonal of the covariance matrix. We emphasize that {\it any} modified kernel discussed in the citations of Section \ref{sec:2}  can be used within our framework and is a focus of our future work.

Once the GP is fit over multi-objective data, acquisition functions, $\mathcal{A}^l$, explore the surrogate model to maximize reward by balancing exploration and exploitation. Using a standard optimization scheme is problematic when dealing with mixed-variable and multi-objective problems due to non-smooth surrogate surface and conflicting objectives. We propose using a constrained, multi-objective GA to optimize the acquisition functions, which, although expensive to use on an actual black-box function, is an ideal candidate for optimizing the acquisition function working on the surrogate surface. For multi-objective problems, multi-objective GA algorithms, such as \cite{996017}, are ideal candidates for obtaining a Pareto-front of optimal solutions. 

Within a GA generation, for multi-objectives, diversification is ensured by a `distance crowding function' which ranks the members of a non-dominated Pareto-front. The `distance crowding function' can be computed in decision-variable space, in function space or a hybrid of the two, and ensures that the generations are distinct and diverse. This inherent feature of GA is exploited to ensure diversity in the `$Q$-batch' of points. The ranking takes place when choosing the test points from an acquisition function for a multi-objective problem  because the choice must take into account the diversity of the solution and propagate the Pareto-front. Because the members of the population are ranked by the GA, we can easily extract a `$Q$-batch' of points from each of the acquisition functions without needing to add any `fantasy' cost function evaluations or optimizing the acquisition functions again. This is a great advantage of using GA as our optimizer since we can output a `$Q$-batch' of diverse query points using the inherent GA features. 

For dealing with mixed-variable problems, GA are again ideal candidates. Genetic algorithms (GA) can be easily be constrained to work in mixed variable spaces. These variables can be dealt with by using probabilistic mutation rates. The genes are allowed to mutate within their prescribed categories, thereby constraining the proposed test points to the $\mathcal{W}$ space.

Common acquisition functions, such as EI, PI, and UCB, can be used within this framework and can be used to nominate a `$Q$-batch' of points. If a candidate in a Q-batch is within the tolerance limit of another candidate in the same batch or a previous data point (for convex functions), we mutate the proposed point within $\mathcal{W}$  to avoid sampling the same data point again.

Test points are selected from $\mathcal{W}$ to evaluate their $\vec{f}$ using HedgeMO algorithm which is detailed in the next section. HedgeMO selects a `$Q$-batch' of test-points from the candidates proposed by each of the acquisition functions. These points are then, along with their function evaluations $\vec{f}$s, appended to the data set.
\subsection{HedgeMO Algorithm}
Hedge strategies use a portfolio of acquisition functions, rather than a single acquisition function. It is an extension to multi-objective problems of GP-Hedge algorithm proposed by \cite{brochu2011portfolio}. 
HedgeMO is part of our MixMOBO algorithm that not only extends the Hedge strategy to multi-objective problems, but also allows `$Q$-batches'. Our algorithm is shown in Algorithm \ref{alg:mohedge}.

\begin{algorithm}[tb]
  \caption{HedgeMO Algorithm}
  \label{alg:mohedge}
\begin{algorithmic}[1]
  \State {\bfseries Input:} Surrogate function $\vec{g}(\vec{w}):{\vec{w} \in \mathcal{W}}$, proposed test points by AFs $\left(\left\{(\vec{u})_{1:n}^{1:L}\right\}_{1:Q}\right)$, batch points per epoch $Q$, current epoch $n$, total objective $K$, parameter $\eta \in \mathbb{R}^{+}$
  \vspace{0.2cm}
  \For{$l=1$ {\bfseries to} $L$}
  \vspace{0.15cm}
  \State For $l^{th}$ acquisition function, find rewards for $Q$-batch points nominated by that AF from epochs ${1{:}n}$-1, by sampling from $\vec{g}$, $\left\{\vec{\theta}^l _ {1:n-1}\right\}_{1:Q} = \vec{\mu}(\left\{(\vec{u})_{1:n-1}^{l}\right\}_{1:Q})$, where $\vec{\theta}=\{\theta\}^k$ for each objective $k$
  \vspace{0.15cm}
  \EndFor
  \vspace{0.2cm}
  \State Normalize rewards for each $l^{th}$ AF and $k^{th}$ objective, $\phi_l^k=\sum_{j=1}^{n-1}\sum_{q=1}^{Q}\frac{\left\{{\theta}^l _ {j}\right\}^k_{q}-min(\Theta)}{max(\Theta)-min(\Theta)}$, where $\Theta$ is defined as $\Theta=\left\{{\theta}^{1:L} _ {1:n-1}\right\}^k_{1:Q}$
  \State Calculate probability for selecting nominees from $l^{th}$ acquisition function, $p^l=\frac{exp(\eta\sum^K_{k=1}\phi_l^k)}{\sum_{i=1}^L exp(\eta\sum^K_{k=1}\phi_i^k)}$
  \vspace{0.2cm}
  \For{$q=1$ {\bfseries to} $Q$}
  \vspace{0.15cm}
  \State Select $q^{th}$ nominee as test-point ${\vec{w}_n}^q$ from $l^{th}$ AF with probability $p^l$
  \vspace{0.15cm}
  \EndFor
  \vspace{0.2cm}
\State {\bfseries return} Batch of test points $\left\{\vec{w}_n\right\}_{1:Q}$
\end{algorithmic}
\end{algorithm}

Extending the methodology presented by \cite{brochu2011portfolio}, HedgeMO chooses the next `$Q$-batch' of test points from the history of the candidates nominated by all of the acquisition functions. Rewards are calculated for each acquisition function from the surrogate surface for the entire history of the nominated points by the $L$ acquisition functions. The rewards are then normalized to scale them to the same range for each objective. This step is fundamentally important because it prevents biasing the probability of any objective. This type of bias, of course, cannot occur in  single objective problems. The rewards for different objectives $k$ are then summed and the probability, $p^l$, of choosing a nominee from a specific acquisition function is calculated using step~6 in Algorithm \ref{alg:mohedge}.
For a `$Q$-batch' of tests points, the test points are chosen $Q$ times. 

\textbf{Regret Bounds:} The regret bounds derived by \citet{brochu2011portfolio} hold for HedgeMO if and only if the Upper Confidence Bound (UCB) acquisition function is a part of the portfolio of acquisition functions. The regret bounds follow from the work of \citet{2012} who derived cumulative regret bounds for UCB. In essence, the cumulative regret in our case 
is bounded by two sublinear terms as for UCB and an additional term which depends on proximity of the chosen point with the test point proposed by UCB. The interested reader is directed to \citet{2012} and \citet{brochu2011portfolio} for a description of the exact regret bounds and their derivation.

\subsection{SMC Acquisition Function}
We introduce a new acquisition function, Stochastic Monte-Carlo (SMC), which for the maximization of an objective, is defined as:
\begin{equation}
SMC \equiv argmax_{\vec{w} \in \mathcal{W}} [\vec{\mu} ({\vec{w}})+r(\vec{w})],
\label{eq:EMCS}
\end{equation}
where $r(\vec{w})$ is sampled from $U(0,2\sigma (\vec{w}))$, and $\vec{\mu} (\vec{w})$ and $\sigma (\vec{w})$ are the mean and standard deviation returned by the GP at $\vec{w}$, respectively. This is equivalent to taking Monte-Carlo samples from a truncated distribution. For categorical and ordinal variable problems, this acqusition function performs well across a range of benchmark tests \cite{sadvours}. We use this acquisition function as part of our portfolio of HedgeMO in the MixMOBO algorithm.

\begin{figure*}[!ht]
  \centering
  \includegraphics[width=\linewidth]{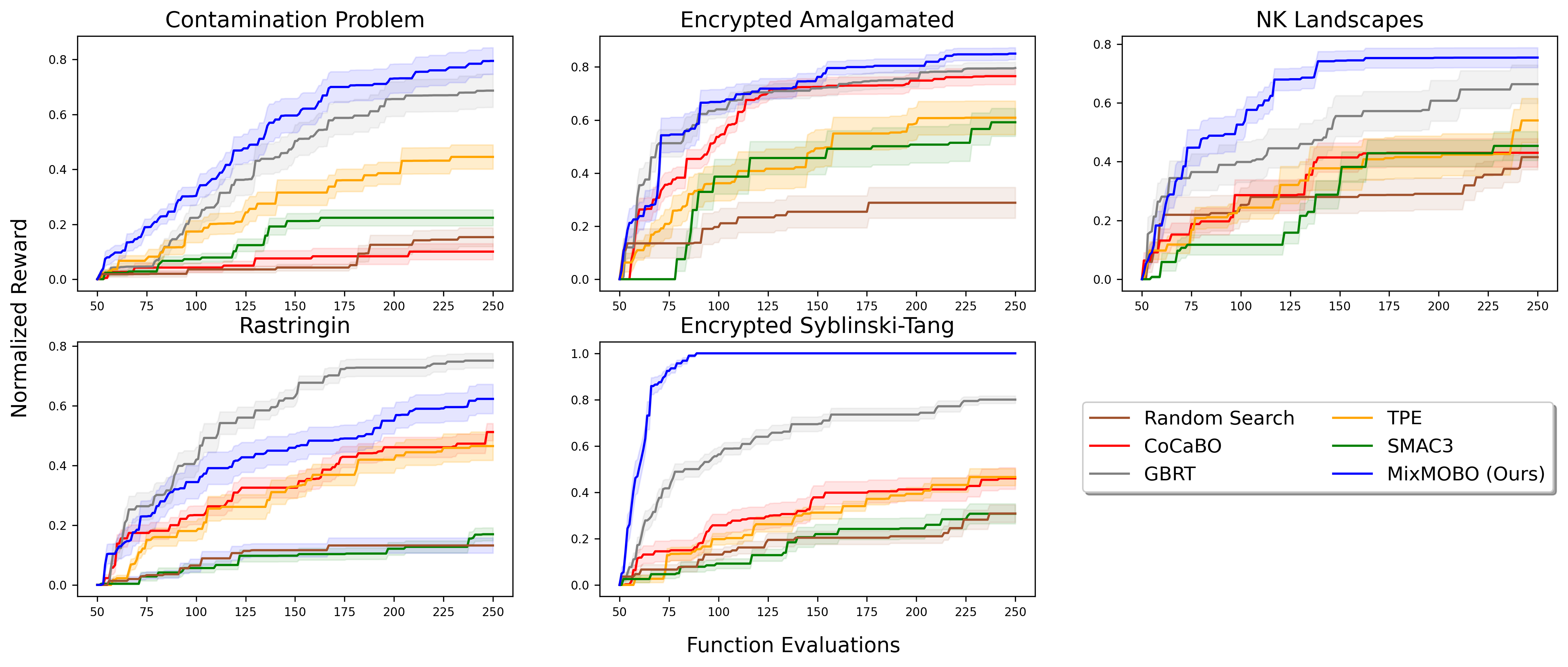}
  \caption{Performance comparison of MixMOBO against other mixed-variable algorithms}
  \label{benchmark}
\end{figure*}

\section{Validation Tests}
\label{VT}
MixMOBO is designed to deal with mixed-variable, multi-objective problems. However, no other small-data algorithm, to the authors' knowledge, can similarly deal with all the attributes of such problems to provide an honest comparison. In the absence of such competition, we use the specific case of mixed-variable, single-objective problems to provide a comparison to state-of-the-art algorithms present for such problems and prove that even for this subset case, MixMOBO is able to perform better than existing algorithms in the literature. We then perform further experiments in both single and multi-objective settings to show the efficacy of the HedgeMO algorithm compared to stand-alone acquisition functions and the performance of SMC in the multi-objective setting.

We benchmarked MixMOBO against a range of existing state-of-the-art optimization strategies that are commonly used for optimizing expensive black-box functions with mixed-variable design spaces. We chose the following single objective optimization algorithms for comparison: \textbf{CoCaBO} \cite{ru2020bayesian}, which  combines the multi-armed bandit (MAB) and Bayesian optimization approaches by using a mixing kernel. CoCaBO has been shown to be more efficient than GPyOpt (one-hot encoding approach \cite{c}) and EXP3BO (multi-armed bandit (MAB approach \cite{NEURIPS2018_cc709032}). We used CoCaBO with a mixing parameter of $0.5$. We also tested MixMOBO against \textbf{GBRT}, a sequential optimization technique using gradient boosted regression trees \cite{skopt}. \textbf{TPE\_Hyperopt} (Tree-structured Parzen Estimator) is a sequential method for optimizing expensive black-box functions,  introduced by \citet{tpe}. \textbf{SMAC3} is a popular Bayesian optimization algorithm in combination with an aggressive racing mechanism \cite{smac3}. Both of these algorithms, in addition to \textbf{Random Sampling}, were used as baselines. Publicly available libraries for these algorithms were used. 

Six different test functions for mixed variables were chosen as benchmarks. A brief description of these test functions and their properties is given below with further details  in Appendix~\ref{app-testfuncs}:

\textbf{Contamination Problem:} This problem, introduced  by \citet{inproceedingscontam}, considers a food supply chain with various stages in the chain where food may be contaminated with pathogens. The objective is to maximize the reward of prevention efforts while making sure the chain does not get contaminated. It is widely used as a benchmark for binary categorical variables. We use the problem as a benchmark with $21$ binary categorical variables.

\textbf{Encrypted Amalgamated:} An anisotropic, mixed-variable function created using a combination of other commonly used test functions \cite{HAL}. We modify the combined function so that it can be used  with mixed variables. In particular, it is adapted for categorical variables by encrypting the input space with a random vector, which produces a random landscape mimicking categorical variables \cite{sadvours}. Our Encrypted Amalgamated function has  $13$ inputs: $8$ categorical, $3$ ordinal variables (with $5$ categories or states each) and $2$ continuous.

\textbf{NK Landscapes:} This is a popular benchmark for simulating categorical variable problems using randomly rugged, interconnected landscapes \cite{KAUFFMAN198711,inproceedingsNKL}. The fitness landscape can be produced with random connectivity and number of optima. The problem is widely used in evolutionary biology and control optimization and is $NP$-complete. The probability of connectivity between $NK$ is controlled by a `ruggedness' parameter, which we set at $20\%$. We test the \citet{inproceedingsNKL} variant with $8$ categorical variables with $4$ categories each.

\textbf{Rastringin:} This is an isotropic test function, commonly used for continuous design spaces \cite{HAL}. We use a $9$-D Rastringin function for testing a design space of $3$ continuous  and $6$ ordinal variables with $5$ discrete states.

\textbf{Encrypted Syblinski-Tang:} This function is isotropic \cite{HAL}, and we  have modified it as we did with the Encrypted Amalgamated test function so that it can work with categorical variables and was used as a representative benchmark for $N$-categorical variable problems. The $10$-D variant tested here consists only of categorical variables with $5$ categories each.

\begin{figure*}
  \centering
  \includegraphics[width=\linewidth]{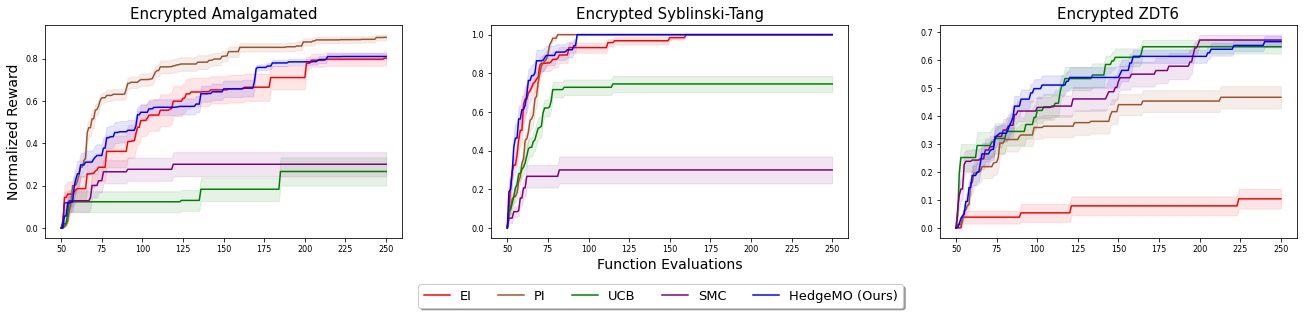}
  \caption{Performance comparison of HedgeMO against other acquisition functions}
  \label{benchmarkAF}
\end{figure*}

\textbf{Encrypted ZDT6:} This is a multi-objective test function introduced by \citet{10.1162/106365600568202} that we modified with encryption so that it can deal with  mixed variables. The test function is non-convex and non-uniform in the parameter space. We test ZDT6 with 10 categorical variables with 5 states each. ZDT6 was only used for testing HedgeMO. 

To the extent of our knowledge, no other optimization algorithm is capable of handling mixed-variable, multi-objective problems in small-data settings. Thus, we have no direct comparisons between MixMOBO and other published algorithms. Therefore, we tested MixMOBO  against a variant of  NSGA-II \cite{996017} with the  ZDT4 and ZDT6 test functions with mixed variables. 
However, we  found that using a GA required more than $10^2$ more  function calls to find the Pareto front to a similar tolerance. For visualization purposes, we do not plot the GA results.

All of the optimization algorithms were run as maximizers, with a $0.005$ noise variance built into all the benchmarks. The budget for each benchmark test was fixed at 250 function calls including the evaluations of 50 initial randomly sampled data points for all algorithms, except for SMAC3 which determines its own initial sample size. The algorithms were run in single output setting (GBRT, CoCaBO and MixMOBO's batch mode was not used for fair comparison). Each algorithm was run 10 times for every benchmark. Our metric for optimization is the `Normalized Reward', defined as \textit{(current optimum - random sampling optimum)/(global optimum - random sampling optimum)}. Figure~\ref{benchmark} shows the Normalized Rewards versus the number of black-box function evaluations for MixMOBO and five other algorithms. The mean and standard deviation of the Normalized Rewards of the $10$ runs for each algorithm, along with their standard deviations (S.D.), are plotted. The width of each of the translucent colored bands is equal to $1/5$ of their S.D.

MixMOBO outperforms all of the other baselines and is significantly better in dealing with mixed-variable problems. GBRT is the next best algorithm and performs better than MixMOBO on the Rastringin function; however, note that the Rastringin function does not include any categorical variables. For problems involving categorical variables, MixMOBO clearly outperforms the others. TPE and CoCaBO have similar performances, and SMAC3 has the poorest performance. All three are outdistanced by MixMOBO.

We then perform experiments to test the performance of our HedgeMO algorithm by comparing it to four different acquisition functions which make up the entirety of its porfolio. These acquisition functions, namely, EI, PI, UCB, and SMC, along with HedgeMO are tested on three different test functions: the Encrypted Amalgamated, Encrypted Syblinski-Tang, 
and Encrypted ZDT6. The latter is used as the multi-objective test function. The Normalized Reward for the multi-objective Encrypted ZDT6 is defined as \textit{(current P-optimum - random sampling P-optimum)/(global P-optimum - random sampling P-optimum)}. Here, \textit{P-optimum}$=\frac{1}{N}\sum_{i=1}^N exp$\textit{(-minimum Hamming distance in parameter space between $i^{th}$ global Pareto-optimal point and any point in the current Pareto-optimal set)}, where $N$ is the number of global Pareto-optimal points.

The results of our acquisition function comparisons are shown in Figure~\ref{benchmarkAF}, which shows that HedgeMO performs well across all three test functions. For single-objective test functions, PI outperforms HedgeMO for Encrypted Amalgamated test function. However, for the multi-objective Encrypted ZDT6 test function, PI performs significantly worse and is outperformed by both SMC and UCB. SMC performs well on multi-objective problems combinatorial problems and hence should be a part of portfolio for a hedging algorithm.

These results prove that for a range of different problems, acquisition function choice can play a huge role in the performance of the algorithm. For a black-box function, this information can not be known a priori, making hedging necessary. HedgeMO consistently performs well in all scenarios and ensures efficiency across a range of different problems. Thus, for unknown black-box functions, HedgeMO should be the hedge strategy of choice for multi-objective problems.

\begin{table*} [!t]
  \caption{Experimental values of the critical buckling $P_c$ (minimization objective for MixMOBO), strain energy density at buckling and fracture, $u_b$ and $u_f$ respectively, elastic stiffness $S$, and ratio of normalized strain energy density compared to the Unblemished structure.}
  \label{materials-table}
  \centering
  \begin{tabular}{cccccc}
    \toprule
    Structure & $P_c [\mu N]$& $u_b [MJm^{-3}]$ & $u_f [MJm^{-3}]$ & $S [MPa]$ & $(u_{fi}/u_{bi})/(u_{f1}/u_{b1})$\\
    \midrule
    Unblemished & 3814.5 &	1.08 &	0.071	& 388.21	& 1\\
    Random Sampling Optimal & 996.2	& 0.08	& 2.85	& 347.19	& 526\\
    MixMOBO Optimal & \bf{545.1}	& \bf{0.02}	& \bf{14.71}	& \bf{460.35}	& \bf{12030}\\

    \bottomrule
  \end{tabular}
\end{table*}

\section{Application to Architected Materials}
\label{AM}
We applied our MixMOBO framework to the optimization of the design of architected, microlattice structures. Advances in modeling, fabrication, and testing of architected materials have promulgated their utility in engineering applications, such as ultralight \cite{zheng2014ultralight,pham2019damage, zhang2019lightweight}, reconfigurable \cite{xia2019electrochemically}, and high-energy-absorption materials \cite{song2019octet}, and in  bio-implants \cite{song2020simple}. The optimization of  architected materials \cite{bauer2016approaching, pham2019damage, xia2019electrochemically, zhang2019lightweight} often requires searching huge combinatorial design spaces, where the evaluation of each design is expensive. \cite{meza2014strong,berger2017mechanical, tancogne20183d}.
\begin{figure}[!htbp]
  \centering
  \includegraphics[width=\linewidth]{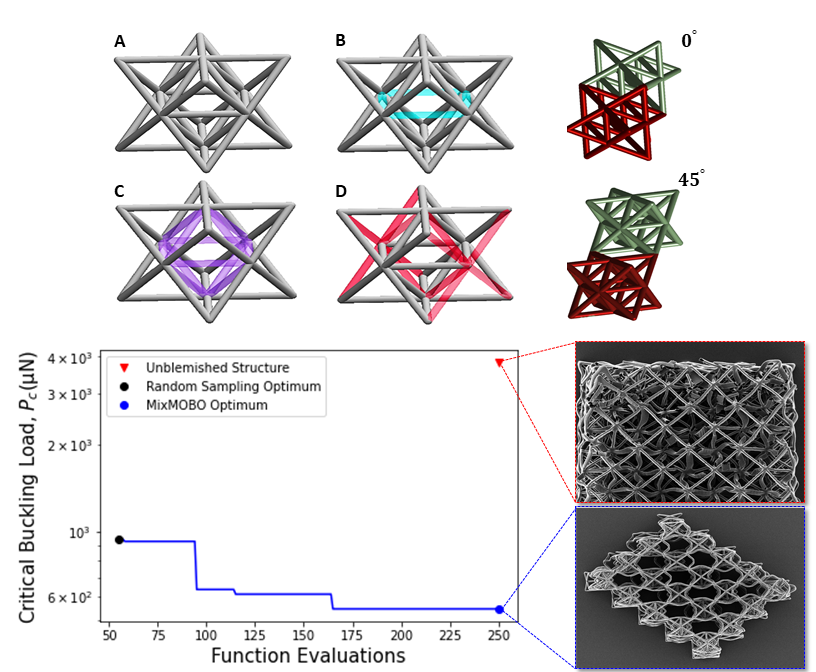}
  \caption{\textit{Top Left:} The 4 unit cells, labelled {\bf A} -- {\bf D}. \textit{Top Right:} The 2 orientations in which they can be joined. \textit{Bottom Left:} Optimization results using MixMOBO. \textit{Bottom Right:} SEM images of Unblemished and Optimum structures.}
  \label{materials}
\end{figure}
The design space for the architected material we optimize here  has $\sim$ 8.5 billion possible combinations of its  17 categorical inputs (one with 2 possible states, and the other 16 with 4 possible states). Our goal is to maximize the strain energy density of a microlattice structure.  We maximize the strain energy density (which is extremely expensive to compute, even for one design) by minimizing the buckling load $P_c$, while maintaining the lattice's structural integrity and stiffness before fracturing. Minimizing $P_c$ (a proxy for maximizing the strain energy density by instigating buckling which leads to the densification of the deformed lattice members) is a more computationally tractable cost function to evaluate (but, it is still expensive and involves solving a numerical finite element code for each evaluation of the cost function.) The manufacturing and testing details of our methodology are included in  \citet{sadvours}, which focuses on the material aspects of the problem.

The design space consists of choosing one of four possible unit cells (shown in the upper left of Fig.~\ref{materials}, each with one or more defects (shown in color) in them, at each of the 16 independent lattice  sites) creating 16 of the categorical inputs with 4 possible values; and the choice of whether the cells are connected along their faces or along their edges on $45^{\text{o}}$-diagonals (shown in the upper right panel of Fig.~\ref{materials}) creating the $17^{th}$ categorical input with 2 possible values.

\begin{figure*}[!ht]
\centering
\includegraphics[trim={4.2cm 0 4cm 0}, scale=.68]{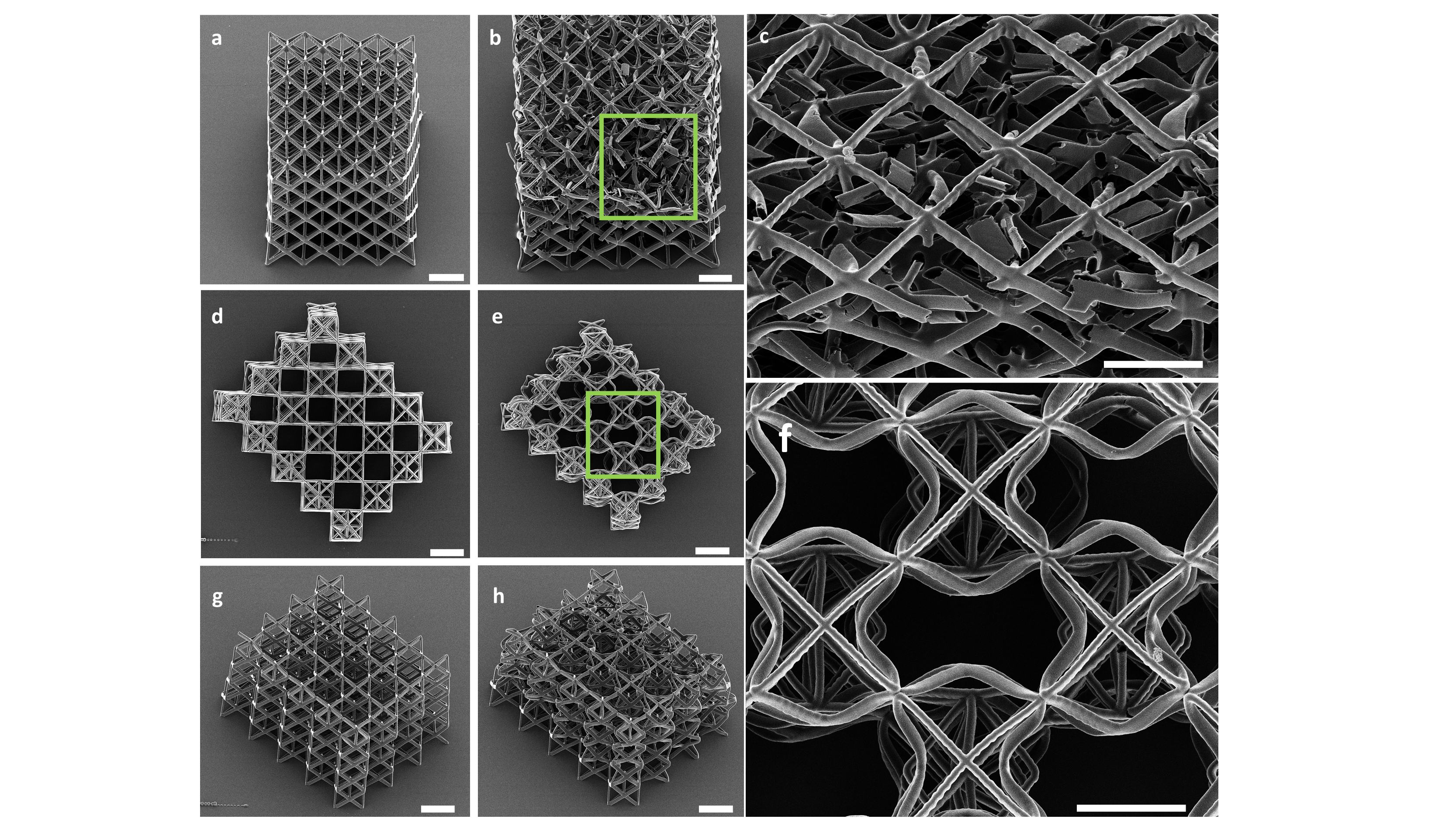}
\caption{{ HIM images of the loaded and unloaded unblemished and optimum structures.} (a)
Unloaded Unblemished structure (b) Unblemished structure after loading, showing severe fracture and collapse of many beam members. (c) Focused image revealing several fractured beams and the internal collapse of the upper layer that subsequently instigated the accumulation of damage in the underlying layers. (d) Unloaded MixMOBO Optimum structure. (e) MixMOBO Optimum structure after the structure was subjected to the same maximum compressive load as the structure shown in (b). Unloading of the optimum structure showed only excessive plastic deformation without catastrophic collapse and the manifestation of the buckling mode. (f) Focused revealing the effect of buckling that led to deformation but no fracture due to the occurrence of densification. (g) Side view of the unloaded optimum structure. (h) Side view of the unloaded optimum structure revealing that fracture was inhibited throughout the structure due to the densification precipitated by the low critical buckling load. \textbf{Scale:} Each scale bar is equal to 10 ${\micro m}$.}
\label{Fig4}
\end{figure*}

The minimization of $P_c$ using MixMOBO was initialized with 50 random structures and the evaluation budget, including initial samples, was set at 250. The algorithm achieved a $42\%$ improvement in the $P_c$ of the lattice structure over the best structure obtained with the first 50 random samples (Figure~\ref{materials}). The optimal microlattice obtained using $P_c$ as a proxy with MixMOBO has an experimentally measured  normalized strain energy density that is {12,030} times greater than that of the unblemished microlattice structure with no defects that is cited in the literature to have the best strain energy density \cite{vangelatos2020regulating}, a $4$ orders of magnitude improvement. Table \ref{materials-table} shows the properties of the fabricated and experimentally measured design created with MixMOBO. The choices of the units cells in the optimally designed lattice that were 
determined by MixMOBO are not intuitive and have no obvious pattern or structure. Images of our optimized structures using Helium Ion Microscopy (HIM) \ref{Fig4}, shows the comparison of the Unblemished structure from literature with our MixMOBO Optimal structure, before and after loading. It is evident that the MixMOBO Optimal structure due to its densification mechanism, can handle much higher loads without breaking \cite{sadvours}.

\section{Conclusions}
The existing optimization literature does not offer an algorithm for optimizing multi-objective, mixed-variable problems with expensive black-box functions. We have introduced  Mixed-variable Multi-Objective Bayesian Optimization (MixMOBO), the first BO based algorithm for optimizing such problems. MixMOBO is agnostic to the underlying kernel. It is compatible with modified kernels and other surrogate methods developed in previous studies for mixed-variable problems. MixMOBO allows for parallel batch updates without repeated evaluations of the surrogate surface, while maintaining diversification within the solution set. We presented the Hedge Multi-Objective (HedgeMO) algorithm, a novel Hedge strategy for which regret bounds hold for multi-objective problems. A new acquisition function, Stochastic Monte-Carlo (SMC) was also proposed as part of the HedgeMO portfolio. MixMOBO and HedgeMO were benchmarked and shown to be significantly better on a variety of test problems compared to existing mixed-variable optimization algorithms. MixMOBO was then applied to the real-world optimization of an architected micro-lattice, and we increased the structure's strain-energy density by $10^4$ compared to existing Unblemished structures in the literature reported to have highest strain energy density. Our future work entails further testing multi-objective and `$Q$-batch' settings. We have also applied MixMOBO for optimization of draft-tubes for hydrokinetic turbines 
\cite{draft} and Cauchy symmetric meta-material structures and are currently applying it for optimization of vertical-axis wind turbines.

\backmatter

\section*{Acknowledgments}
The authors would like to thank Chiyu `Max' Jiang, research scientist at Waymo Research, and Professor Uros Seljak, Department of Physics, University of California at Berkeley (UCB) for insightful discussions regarding Bayesian optimization. We would also like to thank Zacharias Vangelatos and Professor Costas P. Grigoropoulos, Department of Mechanical Engineering, University of California at Berkeley (UCB) for the collaboration to design and manufacture architected materials, conduct nanoindentation, SEM, and HIM experiments. This work used the Extreme Science and Engineering Discovery Environment (XSEDE), which is supported by National Science Foundation grant number ACI-1548562 through allocation TG-CTS190047.




\section*{Declarations}


\subsection*{Funding and Competing Interests}
The authors declare funding sources, conflicts or competing interests that need to be disclosed.

\subsection*{Availability of Data and Materials}
Complete data sets to reproduce any and all experiments which were generated during and analysed during the current study and MixMOBO code are available from the corresponding author on reasonable request.

\subsection*{Authors' Contributions}
H.M.S conceptualized the algorithm, designed the methodology and performed experiments under the supervision of P.S.M. H.M.S and P.S.M then wrote and edited the manuscript.

\section*{Replication of Results}
All the results in this manuscript can be replicated. The complete data sets, MixMOBO algorithm or any other supplementary material and information required for replication are available from the corresponding author on reasonable request.






\bibliographystyle{unsrtnat}
\bibliography{biblio}


\begin{appendices}
\onecolumn
\newpage
\section{Benchmark Test Functions}
\label{app-testfuncs}
In this section, we define the benchmark test functions, all of which are set to be  maximized during our optimizations.
\subsection*{Contamination Problem}
The contamination problem was introduced by \citet{inproceedingscontam} and is used to test categorical variables with binary categories. The problem aims to maximize the reward function for applying a preventative measure to stop contamination in a food supply chain with $D$ stages. At each $i^{th}$ stage, where $i\in[1,D]$, decontamination efforts can be applied. However, this effort comes at a cost $c$ and will decrease the contamination by a random rate $\Gamma_i$. If no prevention effort is taken, the contamination spreads with a rate of $\Omega_i$. At each stage $i$, the fraction of contaminated food is given by the recursive relation: 
\begin{align}
  Z_i=\Omega_i(1-w_i)(1-Z_{i-1}) + (1-\Sigma_i w_i) Z_{i-1}  
\end{align}
here $w_i\in{0,1}$ and is the decision variable to determine if preventative measures are taken at $i^{th}$ stage or not. The goal is to decide which stages $i$ action should be taken to make sure $Z_i$ does not exceed an upper limit $U_i$. $\Omega_i$ and $\Sigma_i$ are determined by a uniform distribution. We consider the problem setup with Langrangian relaxation \cite{baptista2018bayesian}:
\begin{align}
f({\vec{w}})= -\sum_{i=1}^D\left(cw_i+\frac{\rho}{T}\sum_{k=1}^T1_{\{Z_k>U_i\}}\right) - \lambda ||\vec{w}||_1
\end{align}
Here violation of $Z_k<U_i$ is penalized by $\rho=1$ and summing the contaminated stages if the limit is violated and our total stages or dimensions are $D=21$. The cost $c$ is set to be 0.2 and $Z_1=0.01$. As in the setup for \cite{baptista2018bayesian}, we use $T=100$ stages, $U_i=0.1$, $\lambda=0.01$ and $\epsilon=0.05$.

\subsection*{Encrypted Amalgamated}
Analytic test functions  generally cannot mimic mixed variables. 
To map the continuous output of a function into $N$ discrete ordinal or categorical variables, the continuous range of the test function's output is first  discretized into $N$ discrete subranges by selecting $(N-1)$  break points, often equally spaced,  within the bounds of the range. Then, the continuous output variable is assigned the integer round-off value of the subrange defined by its surrounding pair of break  points.  If necessary, the domain of the test function's output is first mapped into a larger domain so that each subrange has a unique integer value. To mimic ordinal variables, we are done, but for categorical variables, a random vector for each categorical variable is then generated which scrambles or `encrypts' the indices of these values, thus creating random landscapes as is the case for categorical variables with a latent space. The optimization algorithm only sees the encrypted space and the random vector is only used when evaluating the black-box function. 

We also define a new test function that we call the {\it Amalgamated function}, a piece-wise function formed from commonly used analytical test functions with different features (for more details on these functions we refer to \citet{HAL}). The Amalgamated function is non-convex and anisotropic, unlike conventional test functions where isotropy can be exploited.

For $i=1...n$, $k=$mod$(i-1,7)$:
\begin{equation}
  f(\vec{w})=\sum_{i=1}^{D}g(w_i)  
\end{equation}
where
\begin{equation}
\begin{aligned}
g(w_i)=
\left\{
\begin{array}{@{\:}l@{}l}
    sin(w_i) & \text{if}\:  k=0, \ w_i\in(0,\pi)\\[1ex]
    -\frac{w_i^4-16w_i^2+5w_i}{2} & \text{if}\:  k=1, \ w_i\in(-5,5)\\[1ex]
    -(w_i^2) & \text{if}\: k=2, \ w_i\in(-10,10)\\[1ex]
    -[10+w_i^2-10cos(2\pi w_i)] & \text{if}\: k=3, \ w_i\in(-5,5)\\[1ex]
    -[100(w_i-w_{i-1}^2)^2+(1-w_i)^2] \quad \qquad & \text{if}\: k=4, \ w_i\in(-2,2)\\[1ex]
    abs(cos(w_i)) & \text{if}\: k=5, \ w_i\in(-\pi/2,\pi/2)\\[1ex]
    - w_i & \text{if}\: k=6, \ w_i\in(-30,30)
\end{array}
\right.
\end{aligned}
\end{equation}\

To create the Encrypted Amalgamated function, for categorical and ordinal variables, equally spaced points are taken within the bounds defined above. For our current work, we use a $D=13$ with $8$ categorical and $3$ ordinal variables with $5$ states each, and $2$ continuous variables.

\subsection*{NK Landscapes}
NK Landscapes were introduced by \citet{KAUFFMAN198711} as a way of creating optimization problems with categorical variables. $N$ describes the number of genes or number of dimensions $D$ and $K$ is the number of epistatic links of each gene to other genes, which describes the `ruggedness' of the landscape. A large number of random landscapes can be created for given $N$ and $K$ values. The global optimum of a generated landscape for experimentation can only be computed through complete enumeration. The landscape cost for any vector is calculated as an average of each component cost. Each component cost is based on the random values generated for the categories, not only by its own alleles, but also by the alleles in the other genes connected through the random epistasis matrix, with $K$ probability or ruggedness. A $K=1$ ruggedness translates to a fully connected genome.

The $NK$ Landscapes from \citet{KAUFFMAN198711} were formulated only for binary variables. They were extended by \citet{inproceedingsNKL} for multi-categorical problems, which is the formulation we use.  Details of the $NK$ Landscape test-functions we use can be found in \citet{inproceedingsNKL}. For the current study, we use $N=8$ with 4 categories each and ruggedness $K=0.2$.

\subsection*{Rastringin} 
Rastringin function is a commonly used non-convex optimization function \cite{HAL} with a large number of local optima. It is defined as:
\begin{equation}
    f(\vec{w})= -[10+w_i^2-10cos(2\pi w_i)], \ w_i\in(-5,5)
\end{equation}
We use $D=9$ for testing with 6 ordinal with 5 discrete states and 3 continuous variables. The ordinal variables are equally spaced within the bounds.

\subsection*{Encrypted Syblinski-Tang}
We use the {\it Syblinski-Tang function} \cite{HAL}, an isotropic non-convex function. The function is considered difficult to optimize because many search algorithms get `stuck' at a local optimum. For use with categorical variables, we encrypt it as described previously. The Syblinski-Tang function, in terms of input vector ${\vec{w}}$, is defined as:
\begin{equation}
    f(\vec{w})= -\frac{\sum_{i=1}^{D}w_i^4-16w_i^2+5w_i}{2}, \ w_i\in(-5,2.5)
\end{equation}
For the current study, this function was tested with $D=10$ categorical variables and 5 categories for each variable.

\subsection*{Encrypted ZDT6}
$ZDT$ benchmarks are a suite of multi-objective problems, suggested by \citet{10.1162/106365600568202}, and most commonly used for testing such problems. We use $ZDT6$, which is non-convex and non-uniform in its parameter space. We again modify the function by encrypting it to work with categorical problems. $ZDT6$ is defined as:
\begin{equation}\
\begin{aligned}
 f_1(\vec{w})&=exp(-4w_1)sin^6(6 \pi w_1)-1 \\ f_2(\vec{w})&=-g(\vec{w})\left[1-(f_1(\vec{w})/g(\vec{w}))^{2}\right]\\ g(\vec{w})&=1+9\left[\left(\sum_{i=2}^Dw_i\right)/(n-1)\right]^{1/4} 
\end{aligned}
\end{equation}\
 Here $ w_1 \in [0,1]$ and $ w_i =0$ for $i = 2,\dots,D$. The function was tested for $D=10$ with 5 categories each. We note that to evaluate the performance of MixMOBO, we compared it against the NSGA-II variant \cite{996017} that can deal with mixed variables (by running $ZDT4$ in a mixed variable setting and $ZDT6$ with categorical variables). No encryption is necessary for GAs. GAs required, on average, $10^2$ more function calls compared to MixMOBO.

\end{appendices}

\end{document}